Original Article

# Learning to control and coordinate mixed traffic through robot vehicles at complex and unsignalized intersections

Dawei Wang[1], Weizi Li[2], Lei Zhu[3] and Jia Pan[1]

## Abstract

*Intersections are essential road infrastructures for traffic in modern metropolises. However, they can also be the bottleneck of traffic flows as a result of traffic incidents or the absence of traffic coordination mechanisms such as traffic lights. Recently, various control and coordination mechanisms that are beyond traditional control methods have been proposed to improve the efficiency of intersection traffic by leveraging the ability of autonomous vehicles. Among these methods, the control of foreseeable mixed traffic that consists of human-driven vehicles (HVs) and robot vehicles (RVs) has emerged. We propose a decentralized multi-agent reinforcement learning approach for the control and coordination of mixed traffic by RVs at real-world, complex intersections—an open challenge to date. We design comprehensive experiments to evaluate the effectiveness, robustness, generalizability, and adaptability of our approach. In particular, our method can prevent congestion formation via merely 5% RVs under a real-world traffic demand of 700 vehicles per hour. In contrast, without RVs, congestion will form when the traffic demand reaches as low as 200 vehicles per hour. Moreover, when the RV penetration rate exceeds 60%, our method starts to outperform traffic signal control in terms of the average waiting time of all vehicles. Our method is not only robust against blackout events, sudden RV percentage drops, and V2V communication error, but also enjoys excellent generalizability, evidenced by its successful deployment in five unseen intersections. Lastly, our method performs well under various traffic rules, demonstrating its adaptability to diverse scenarios. Videos and code of our work are available at https://sites.google.com/view/mixedtrafficcontrol.*

## 1. Introduction

Traffic flow is the beating heart of a city, driving economic growth and ensuring daily lives. Despite the implementation of various traffic control methods, including traffic signals, ramp meters, and tolls, traffic congestion continues to be a global issue, with external expenses amounting to over $100 billion annually (Schrank et al., 2021). Modern urban road networks largely consist of linearly-coupled roads interconnected by intersections. The key to this design's functionality is the *intersection*, which enables traffic flows to interchange and disperse. Any intersection blockage can disrupt traffic from all directions, leading to traffic spillover and even city-wide gridlock. Unfortunately, intersections are vulnerable to traffic incidents with more than 45% of all crashes taking place at intersections in the U.S. (Choi, 2010) and are susceptible to extreme weather and energy shortages, which can leave intersections without control for days or even weeks, paralyzing the traffic (Press, 2022; Winck, 2022; Ramirez, 2022). This

raises the question: *How can we ensure uninterrupted traffic flows at intersections?*

While current transport control methods have limited effectiveness in mitigating traffic delays and congestion, connected and autonomous vehicles (CAVs) (Spielberg et al., 2019; Feng et al., 2023; Pek et al., 2020) offer new opportunities. Recent studies (Sharon and Stone, 2017; Yang and Oguchi, 2020) have demonstrated the possibilities of using autonomous vehicles to enhance intersection traffic

[1]Department of Computer Science, The University of Hong Kong, Hong Kong, China
[2]Department of Electrical Engineering and Computer Science, The University of Tennessee, Knoxville, TN, USA
[3]Department of Industrial and Systems Engineering, The University of North Carolina at Charlotte, Charlotte, NC, USA

**Corresponding author:**
Jia Pan, Department of Computer Science, The University of Hong Kong, Rm 410 Chow Yei Ching Building, Hong Kong, China.
Email: jpan@cs.hku.hk

throughput. However, these studies presume universal connectivity and centralized control of all autonomous vehicles, a scenario that may not materialize soon. The transition to varying levels of autonomous vehicles will be *gradual*, with a prolonged period of mixed traffic comprised of both human-driven vehicles (HVs) and robot vehicles (RVs). Despite the challenge in modeling and controlling mixed traffic due to the diversity and suboptimality of human drivers, mixed traffic control is possible through algorithmically determining the behaviors of RVs to regulate HVs (Wu et al., 2022). While progress has been made (see Section 2 for details), no evidence exists to demonstrate the feasibility of controlling mixed traffic through RVs at *real-world, complex intersections* where a large number of vehicles may potentially conflict. However, being able to control traffic at real-world intersections is an essential step toward citywide traffic control and unveiling full societal benefits of autonomous vehicles (Urmson and Whittaker, 2008).

In this project, we study mixed traffic control at real-world intersections through RVs. The intersection layouts and reconstructed traffic are shown in Figure 1 LEFT. To test the limit of mixed traffic control and explore the envisioned benefits of RVs to our traffic system, we further assume these intersections are *unsignalized*, with the flow of traffic entirely and solely controlled by RVs. Numerous challenges abound under such an assumption, such as modeling mixed traffic behavior and designing a representation of traffic conditions that encompasses diverse topologies and dynamically fluctuating real-world traffic demands.

We propose a model-free reinforcement learning (RL) approach for mixed traffic control at complex intersections. When approaching an intersection, an RV will first observe the traffic condition via local perception and vehicle-to-vehicle (V2V) communication, and then encode the observation as input to a mixed traffic control policy. The policy will output a high-level decision of whether the RV should enter or not enter the intersection independently from other RVs. Our approach falls into the paradigm of centralized training and decentralized execution. RVs are centrally trained with a shared policy and reward function, which accounts for traffic efficiency and potential conflicts. During execution, all RVs make independent decisions while collectively ensuring smooth traffic flow at the intersection *without explicit coordination*. We also design a conflict resolution mechanism for eliminating potential conflicts at the intersection, which significantly boosts training efficiency and road safety.

We conduct comprehensive experiments under high-fidelity traffic reconstruction and simulation. The real-world traffic data provided by the city of Colorado Spring, CO, USA is used to reconstruct the simulated traffic flow, validating that our training environments and evaluation experiments closely resemble real-world conditions. Our overall results show that, with 60% or more RVs, our method outperforms traditional traffic light control in terms of traffic efficiency in most scenarios. For example, the average waiting time of all vehicles is reduced by 25.9% and 40.7% compared to employing traffic lights at intersection I, when the RV penetration rate is 70% and 90%, respectively. With 100% RVs, our method reduces the average waiting time of the entire intersection traffic up to 42% compared to traffic light control and 89% compared to the traffic light absence baseline. We further explore the relationship between traffic demands, congestion, and RV penetration rates. We find with just 5% RVs, our method can prevent congestion from developing under the actual traffic demand of 700 vehicles per hour (v/h). In contrast, without RVs, congestion will form at an (unsignalized) intersection when traffic demand reaches as low as 200 v/h.

Besides effectiveness, we also desire robustness, generalizability, and adaptability of our method. For robustness, we conduct a "blackout" experiment to show the ability of our approach to stabilize the traffic flow when traffic lights suddenly stop working and traffic control transitions to our system. During the blackout, the RVs act as self-organized "traffic lights" coordinating the traffic at a high throughput. We also examine the impact of sudden RV rate drops. The results show that even with 40% drop (from 90% to 50%), our method still maintains stable and efficient traffic flows at the intersection. Next, we analyze the impact of observation errors on traffic conditions due to varied V2V connectivities and range settings, simulated through multi-hop communication and packet error events. As a result, our method is robust even under extreme situations, such as 3-hop communication and a 20% packet error rate (PER). This hints the practicality of our method in real-world deployment. For generalizability, we deploy our method without any modification or tuning at five *unseen* intersections: Not only does our method eliminate congestion, but with 60–70% RVs, it also surpasses traffic light control in reducing the average waiting time for all vehicles. For adaptability, our method can adapt to various traffic rules, such as left-hand traffic, by incorporating and regulating the right-turn traffic streams. The results show comparable performance to the original policy, illustrating the flexibility of our method and its applicability across different countries.

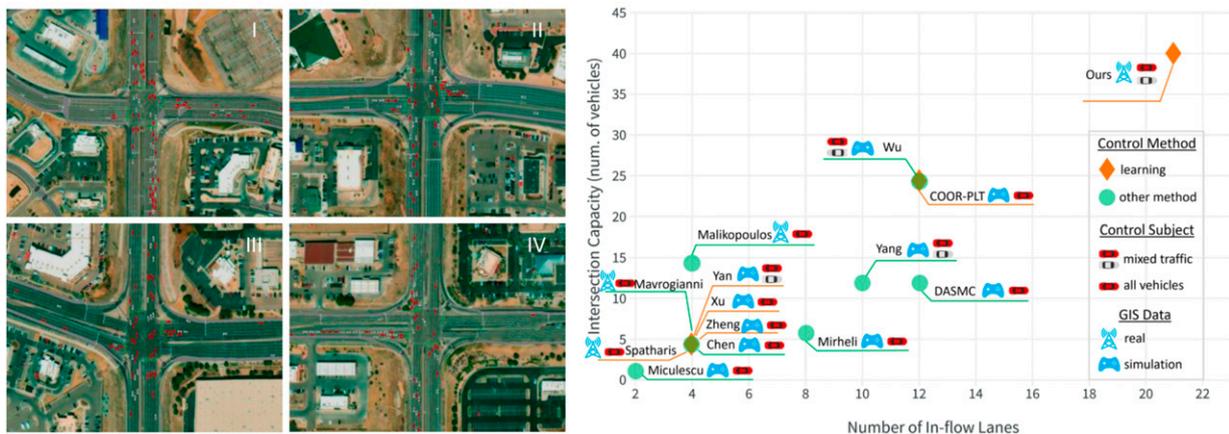

**Figure 1.** LEFT: We study real-world, complex intersections located at Colorado Springs, CO, USA. The traffic is reconstructed using the actual traffic data collected at these intersections (more details of these intersections are in Table 2). RIGHT: Comparison of state-of-the-art studies on intersection traffic control. The references can be found in Section 2 related work.

In summary, our work is the first to demonstrate the feasibility of controlling and coordinating mixed traffic at unsignalized intersections with complex topologies and real-world traffic demands. As many challenges are addressed for the first time in mixed traffic control, we hope that our design can provide insights into these challenges and stimulate future endeavors in the field. Our code can be found at https://github.com/daweidavidwang/MixedTrafficControl.

## 2. Related work

### 2.1. Unsignalized intersection control

There are two common approaches to control and coordinate traffic at intersections (Rios-Torres and Malikopoulos, 2016). The first and extensively studied is traffic signal control (Wei et al., 2019; Mohamed and Radwan, 2022; Jácome et al., 2018). Our work differs from this line of research by considering intersections unsignalized. The second is the intersection management system (Miculescu and Karaman, 2019; Malikopoulos et al., 2018), which commonly requires the central control of all vehicles. This is not applicable to mixed traffic given the flexibility of HVs. Recently, there has been growing research on using reinforcement learning (RL) to control and coordinate traffic due to RL's model-free nature and its ability to test a diverse range of scenarios through simulation without jeopardizing the safety of participants (Yan and Wu, 2021; Yan et al., 2022).

We analyze the complexity of scenarios utilized in previous methods for controlling unsignalized intersections. The comparison of our work and example studies is presented in Figure 1 RIGHT. As these studies do not provide all measurements, we offer our best estimates. To provide some details, COOR-PLT (Li et al., 2023a) uses a two-layer hierarchical RL approach that centrally forms RV platoons, and then decentrally coordinate those platoons at the intersection. DASMC (Zhou et al., 2022) combines microscropic-level virtual platooning and macroscopic-level traffic flow regulation to coordinate RVs across unsignalized intersections. Yang and Oguchi (2020) propose an intersection delay model that predicts total vehicle delay and assigns optimally-controlled actions to RVs to minimize the delay. Malikopoulos et al. (2018) define the control of intersection traffic as an energy minimization program. By solving the program, lower fuel consumption and travel times are achieved. Mirheli et al. (2019) define a cooperative trajectory planning to control and coordinate RVs through unsignalized intersections. Chen et al. (2022) consider local conflicts among RVs at unsignalized intersections. Yan and Wu (2021) use RL to control mixed traffic through RVs at two-way/four-way unsignalized intersections. Miculescu and Karaman (2019) define a polling systems-based algorithm that provides safe and efficient coordination. Spatharis and Blekas (2024) present a multi-agent RL method to control traffic at intersections. Xu et al. (2021) introduce a centralized scheme for scheduling autonomous vehicles under signal-free conditions. Zheng et al. (2022) propose a cooperative multi-agent proximal optimization algorithm for controlling connected autonomous vehicles. Mavrogiannis et al. (2023) present a method for abstracting road traffic to aid in traffic analysis and vehicle control at intersections. Wu et al. (2023) achieve unsignalized intersection control using mixed integer nonlinear programming. Yan et al. (2021) propose an RL method to optimize mixed traffic flow at three-way intersections. Our work differentiates from all studies mentioned above, as we specifically address intersection scenarios with complex topologies and real-world traffic demands—an open challenge to date.

### 2.2. Mixed traffic control

Traditional mathematical approaches, such as defining and solving an optimization or control problem, are common solutions presented for mixed traffic control problems (Wang et al., 2019; Karimi et al., 2020; Cai et al., 2020; Dai et al., 2021; Wang et al., 2023b; Lu et al., 2023; Hickert et al., 2023). For example, Yang and Oguchi (2020) solve an optimization problem for controlling and coordinating mixed traffic through an unsignalized intersection, while work by Zhao et al. (2018) solve an optimization problem for coordinating mixed traffic at roundabouts. Wu et al. (2018) introduce an optimal linear controller to stabilize traffic flow on freeways. However, there are issues with solving mixed traffic control problems through traditional approaches as they typically require explicit modeling of the system's traffic flow or do not holistically capture the underlying traffic dynamics. As such, recent studies explore using RL as an alternative, given RL's ability to handle mixed traffic's complex behaviors without making the same traffic flow/dynamic assumptions.

Recent studies have demonstrated the potential of mixed traffic control via RL in scenarios such as ring roads, figure-eight roads (Wu et al., 2022; Poudel et al., 2024), highway bottleneck and merge (Vinitsky et al., 2018; Feng et al., 2021), two-way intersections (Yan and Wu, 2021; Villarreal et al., 2024), and roundabouts (Jang et al., 2019; Chinchali et al., 2019). However, these scenarios typically lack real-world complexity and only involve a limited number of conflicting vehicles, which contrast to our work where we ensure both are present.

## 3. Methodology

The pipeline of our approach is shown in Figure 2. Each RV entering the control zone employs our method and observes the traffic condition within the zone. The RV subsequently encodes the traffic condition into a fixed-length representation (as shown in Figure 2(a) and elaborated in Section 3.2.2), and then uses it to make a high-level decision (Stop or Go) at the intersection entrance (as shown in Figure 2(b) and elaborated in Section 3.2.1).

### 3.1. Intersection traffic

A standard four-way intersection comprises four moving directions: eastbound (E), westbound (W), northbound (N), and southbound (S); and three turning options: left (L), right (R), and cross (C). As an example, we use E-L and E-C to denote left-turning traffic and crossing traffic that travel eastbound, respectively. The complete notation is shown in Figure 2(a). We further define "conflict" as two moving directions intersecting each other, for example, E-C and N-C. In most right-hand driving countries, such as the U.S. and China, right-turning vehicles are typically not required to wait for the green light. Hence, it is less important to coordinate right-turning traffic, which has minimal impact on intersection traffic flow. To accommodate this observation, we consider eight traffic streams that may lead to conflicts: E-L, E-C, W-L, W-C, N-L, N-C, S-L, and S-C; and we define the conflict-free set $\mathcal{C}$ as (S-C, N-C), (W-C, E-C), (S-L, N-L), (E-L, W-L), (S-C, S-L), (E-C, E-L), (N-C, N-L), and (W-C, W-L). Conflicts may arise for the pairs of traffic streams that are not in $\mathcal{C}$.

Our design retains flexibility to accommodate various traffic rules, including left-hand driving countries, or instances where both right-turn and left-turn directions require control. The definition of the eight traffic streams and conflict-free set allow seamless adherence to a specific traffic regulation. To demonstrate that, we extend our approach to control 12 moving directions at four-way intersections. The experiments in Section 4.11 show that the 12-direction policy achieves similar performance to our baseline policy.

### 3.2. Decentralized RL for mixed traffic

We formulate mixed traffic control as a partially observable Markov decision process (POMDP), which consists of a seven-tuple $(\mathcal{S}, \mathcal{A}, \mathcal{T}, \mathcal{R}, \Omega, \mathcal{O}, \gamma)$, where $\mathcal{S}$ is a set of states ($\mathbf{s} \in \mathcal{S}$), $\mathcal{A}$ is a set of actions ($\mathbf{a} \in \mathcal{A}$), $\mathcal{T}$ is the transition probabilities between states $\mathcal{T}(\mathbf{s}' \mid \mathbf{s}, \mathbf{a})$, $\mathcal{R}$ is the reward function ($\mathcal{S} \times \mathcal{A} \rightarrow \mathbb{R}$), $\Omega$ is a set of observations $\mathbf{o} \in \Omega$, $\mathcal{O}$ is the set of partial observations, and $\gamma \in [0, 1)$ is a discount factor. At each time $t$, when the $i$th RV enters the control zone, its action $a_i^t$ is determined based on the current traffic condition $o_i^t$, which is a partial observation of the traffic state $s_i^t$ of the intersection. We present the policy $\pi_\theta$ as a neural network trained using the following loss:

$$\left( R_{t+1} + \gamma_{t+1} q_{\overline{\theta}} \left( O_{t+1}, \arg\max_{a'} q_\theta(O_{t+1}, a') \right) - q_\theta(O_t, A_t) \right)^2, \quad (1)$$

where $q$ denotes the output from the value network; $\theta$ and $\overline{\theta}$ respectively represent the value network and the target

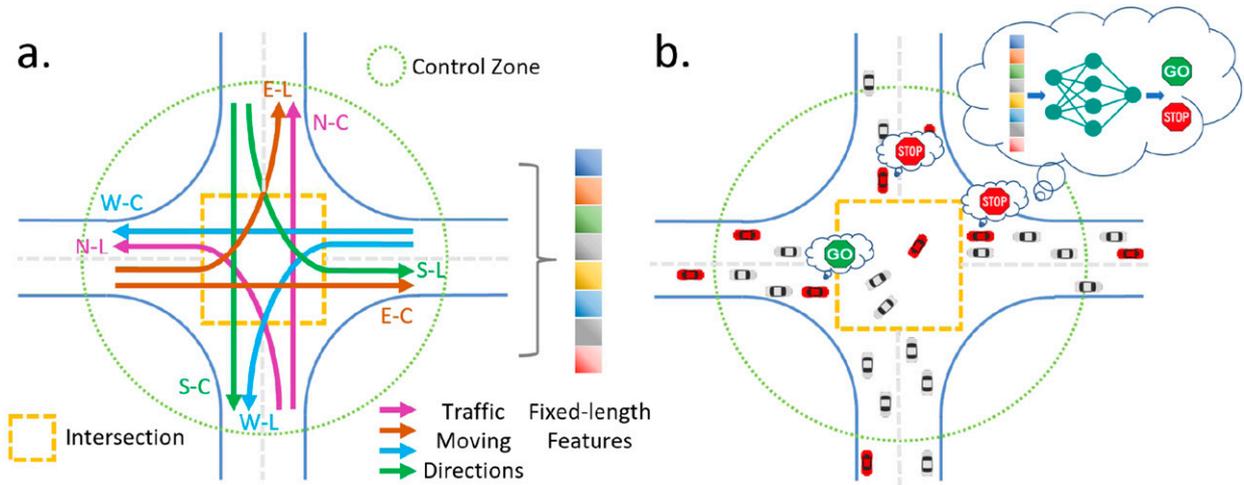

**Figure 2.** The pipeline of our approach. (a) Each RV within the control zone encodes the intersection traffic condition as a fixed-length representation, including both macroscopic traffic features such as queue length and microscopic traffic features such as vehicle locations. E, W, N, and S represent east, west, north, and south, respectively; C means crossing and L means left-turning. (b) The encoded traffic condition is then used by each RV to decide Stop or Go at the intersection entrance to manage mixed traffic.

network (Hessel et al., 2018). The target network is a periodic copy of the value network.

### 3.2.1. Action space.

As our focus is exploring mixed traffic control via RVs over traffic lights, we restrict the action space of RV to high-level decisions $A = \{Stop, Go\}$. An RV's action $a_i^t \in A$ determines whether the RV $i$ should enter the intersection or stop at the intersection entrance to hold its following vehicles.

The longitudinal acceleration of an RV is computed using intelligent driver model (IDM) (Treiber et al., 2000) when the vehicle is outside the control zone. Within the control zone, if the RV decides Go, it accelerates using the maximum acceleration $a^t = a_{max}$; conversely, if the RV decides Stop, it decelerates and comes to a halt via $a^t = -v^2/2d_{front}$, where $d_{front}$ is the distance to the intersection. In the event of a potential collision, the emergency brake is automatically engaged via the built-in collision avoidance mechanism (Krauss, 1998) of the RV, overriding the requested acceleration. Further discussion of the RV is provided in Section 3.4.

### 3.2.2. Observation space.

To empower an RL policy to generalize across diverse intersection topologies, we encode the traffic condition observed by each RV into a fixed-length representation. The observation for each RV within the control zone (commencing from 30 m before the intersection) has three elements.

- The status of RV: The distance from RV $i$'s current position to the intersection, denoted as $d_i^t$.
- Traffic conditions within the control zone but outside the intersection: The queue length $l^{t,j}$ and the average waiting time $w^{t,j}$ of each of the eight traffic moving directions that are defined in Section 3.1. These features quantify the anisotropic congestion levels of an intersection. In simulation, $l^{t,j}$ is computed as the number of vehicles before the last RV in the control zone along direction $j$, and $w^{t,j}$ is computed as the average waiting time of RVs in the control zone along direction $j$. Note that the values of $l^{t,j}$ and $w^{t,j}$ can be smaller than the actual values when RV's penetration rate is low. In real world, these features can be estimated by each RV through V2V communication (Cheng et al., 2015). More discussion can be found in the next section.
- Traffic condition inside the intersection: We design an occupancy map $m^{t,j}$ for each moving direction. As depicted in Figure 3 LEFT, we divide the inner lane along a moving direction into 10 equally-sized segments. A segment is labeled "occupied" with 1 if a vehicle's position is located within it, or labeled "free" with 0 if otherwise. This information is observed through the RV's local perception system (Li et al., 2023b).

Overall, the observation space of RV $i$ at $t$ is

$$o_i^t = \oplus_j^J \langle l^{t,j}, w^{t,j} \rangle \oplus_j^J \langle m^{t,j} \rangle \oplus \langle d_i^t \rangle, \quad (2)$$

where $\oplus$ is the concatenation operator and $J = 8$ is the total number of traffic moving directions.

### 3.2.3. Traffic condition estimation.

To obtain the traffic condition outside the intersection (but within the control zone), we use a decentralized approach: the $i$th RV stops inside the control zone, it will compute its waiting time $^{ego}w_i^t$ and the queue length of its direction $^{ego}l_i^t \approx d_i^t/v_{len}$. Here, $d_i^t$ represents the distance from the current position of the $i$th RV to the intersection, and $v_{len}$ is the vehicle length plus the distance between cars, which is set to 5 m. The waiting time and queue length are propagated to all RVs in the control zone and aggregated to form the overall traffic condition along each moving direction.

The traffic condition estimation can be subject to multiple errors. First, the uniform 5 m vehicle length may lead to inaccurate estimations of the queue length. Second, only RVs within the control zone broadcast their observations. Therefore, if most vehicles inside the control zone are HVs, $w^{t,j}$ and $l^{t,j}$ may be underestimated. Third, the lack of the contribution of RVs outside the control zone can result in estimation bias. Lastly, communication issues such as poor wireless connectivity can deteriorate the estimation results. Thanks to our reward design and conflict resolution mechanism, our method is robust against these issues. See details in Section 4.10.

### 3.2.4. Conflict-aware reward.

To encourage the RV to consider not only its own efficiency but also the conflicts within the intersection, we design a conflict-aware reward function for the RV:

$$r(s^t, a^t, s^{t+1}) = \lambda_L r_L + p_c, \quad (3)$$

where $r_L$ is the local reward, $p_c$ is the conflict punishment, and $\lambda_L$ is the coefficient.

The local reward $r_L$

$$\begin{cases} -w^{t+1,j}, & \text{if } a^t = \text{Stop}; \\ w^{t+1,j}, & \text{otherwise.} \end{cases} \quad (4)$$

$w^{t+1,j}$ is the average waiting time of all vehicles in the $j$th direction, which is normalized using $w^{t+1,j}/w_{max}$ and $w_{max} = 200$. $p_c$ denotes the punishment for conflicts. If the RV decides Stop, the local reward is the negative waiting time $-w^{t+1,j}$; otherwise, it is positive $w^{t+1,j}$.

The conflict punishment $p_c$ is

$$\begin{cases} -1, & \text{if conflict}; \\ 0, & \text{otherwise.} \end{cases} \quad (5)$$

If the RV's movement conflicts with other vehicles in the intersection, it incurs a penalty of $-1$.

The reward design is inspired by the observation that waiting time (Zhang et al., 2020; Gregurić et al., 2020) has been a popular choice in measuring traffic congestion. We demonstrate the effectiveness of our reward design in improving intersection traffic through extensive experimentation in Section 4.8.

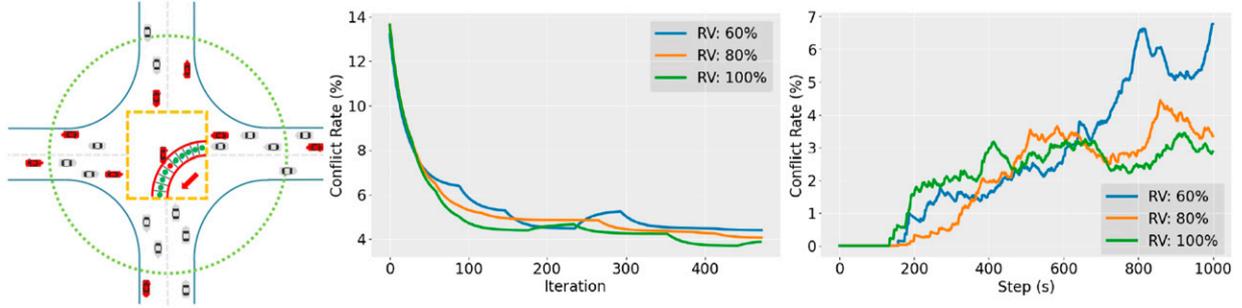

**Figure 3.** LEFT: The occupancy map along the moving direction W-L. Each of the 10 segments is labeled with either free (green dot) or occupied (red dot). MIDDLE: As learning progresses, the frequency of conflicting decisions decreases and stabilizes at a low level over all three RV penetration rates. RIGHT: Regardless of the RV penetration rate, the conflict rate (calculated as the number of conflict decisions divided by the total number of RVs' decisions) stays low, for example, ~6% for 60%, <4% for 80% RVs, and <3% for 100% RVs.

*3.2.5. RL algorithm.* We employ Rainbow DQN (Hessel et al., 2018), a state-of-the-art RL algorithm for discrete action tasks. It combines six extensions of the original DQN algorithm (Mnih et al., 2015). We equip it with our reward function to centrally train all RVs. During execution, each RV makes independent decisions using the shared policy, structured as a neural network with three fully connected (FC) layers and each FC layer contains 512 hidden units with ReLU as the activation function. Training takes around 48 h using Intel i9-13900K and NVIDIA GeForce RTX 4090. Other hyperparameters are listed in Table 1.

We choose Rainbow DQN for its outstanding performance on discrete action tasks. However, we have explored other RL algorithms for comparison. As shown in Figure 19, both Rainbow DQN (Hessel et al., 2018) and PPO (Schulman et al., 2017) yield similar results, while SAC (Haarnoja et al., 2018) lags slightly. This result shows that our novelty and contribution are not hinged on a specific RL algorithm.

### 3.3. Conflict resolution mechanism

The fundamental cause of intersection congestion and accidents is the conflicting directions of movement. Although we penalize RVs for conflicting decisions, our reward function may not completely eliminate conflicts, that is, Go decisions of RVs from conflicting directions.

Effective learning can only take place if less conflicts occur during training: Conflicts will lead to congestion, which further hinders sampling and training. To avoid congestion, we incorporate a conflict resolution mechanism to post-process the RL outputs. If there are no vehicles on conflicting streams or inside the intersection, and no conflicting decisions among the RVs, an ego RV who decides Go will enter the intersection. If there are vehicles inside the intersection, particularly on the conflicting streams of the ego RV, the ego RV is not permitted to enter the intersection. When multiple RVs on conflicting streams arrive at the intersection entrance and all decide Go, the RV with the highest priority score (calculated by averaging waiting time and queue length) is granted entry, while the others must wait.

**Table 1.** Hyperparameters of our RL algorithm.

| Parameters | Value |
| --- | --- |
| $\lambda_L$ | 1 |
| Acceleration | 2.6 m/s² |
| Deceleration | −4.5 m/s² |
| Prioritized replay buffer $\alpha$ | 0.5 |
| Replay buffer capacity | 50,000 |
| Number of atoms | 51 |
| Hidden layers | [512, 512, 512] |
| Discount factor | 0.99 |
| Minibatch size | 32 |
| Learning rate | 0.0005 |
| Control zone radius | 30 m |

We evaluate the effectiveness of the conflict resolution mechanism in Section 4.9. The results not only demonstrate the effectiveness of our approach, but also justify the necessity of the mechanism.

### 3.4. Assumptions of robot vehicles

Our method focuses on high-level decisions (Stop/Go) and requires only basic V2V communication to obtain the positions and decisions of other RVs. It can be integrated into autonomous driving software, comprising other modules such as perception, planning, and control to achieve the full self-driving capability. Within the pipeline of self-driving software, various components bear distinct responsibilities for overall safety (Muhammad et al., 2020; Shao et al., 2023). High-level modules positioned upstream, such as perception and decision-making, are not directly accountable for safety outcomes in most self-driving vehicles to date. Downstream modules like planning and control (PnC), on the other hand, assume greater responsibility as they need to navigate the vehicle through environmental uncertainties and ensure collision-free trajectories. We prioritize safety by minimizing potential conflicts within the intersection, thereby inherently enhancing the safety of the entire system. Additionally, our policy is not designed to handle physical collisions as it does not control the vehicle's acceleration. When facing such an event, the PnC module should engage emergency braking to avoid a collision.

However, despite advancements, accidents involving autonomous vehicles in the U.S. have led to lingering doubts about their reliability. To navigate the landscape and tap into the L2/L3 dominated market, our method can serve as a plugin for L2/L3 driving software, acting as a suggestive component. Human drivers retain responsibility for the vehicle's safety, with the option to override suggestions or assume direct control. As demonstrated in Section 4.5, our method exhibits robust coordination of mixed traffic even when some RVs revert to HVs. These features make our method adaptable and practical across all levels of vehicle autonomy in mixed traffic.

For low-level control of the RV, we employ the traffic simulator simulation of urban mobility (SUMO) (Behrisch et al., 2011). It includes human driving models, configurable traffic networks and flows, and mechanisms for enforcing traffic rules, safety rules, and physical constraints. The built-in collision avoidance mechanism (Krauss, 1998) and human driving model (Treiber et al., 2000) will act as the downstream modules of self-driving software and the human driver, respectively. These mechanisms and models will ensure collision-free driving of a vehicle. This assumption has been widely adopted by previous studies (Yan and Wu, 2021; Wu et al., 2022; Zhang et al., 2023; Cui et al., 2021).

## 4. Experiments and results

### 4.1. Mixed traffic

*4.1.1. Reconstruction and simulation.* In order for RVs to interact with HVs under real-world traffic conditions, we need to first reconstruct traffic using actual traffic data and then pursue high-fidelity simulations. We reconstruct the intersection traffic using turning count data at each intersection provided by the city of Colorado Springs, CO, USA.[1] The turning count data records the number of vehicles moving in a particular direction at the intersection and is collected via in-road sensors such as infrastructure-mounted radars. We have in total six intersections' data and we label these intersections I, II, III, IV, V, and VI, respectively. We use intersections I–IV to train the RL policy. Given the GIS data (traffic data and digital map), we pursue traffic simulations in SUMO. A directed graph is used to describe the simulation area: Each edge of the graph represents a road segment with an ID and a vehicle's route is defined by a list of edge IDs.

Vehicles are routed using jtcrouter[2] based on the turning count data. By default, jtcrouter will select edges that are close to the intersection as the starting and ending edges of a route. This can result in extremely short routes and affect the simulation fidelity. To mitigate this issue, we adjust vehicle routes by proposing more suitable edges for vehicles' arrival and departure on the network. Specifically, for traffic streams on the main road that connects the four intersections used in training the RL policy, we assign the starting and ending edges of a vehicle to be the boundary of the main road. For traffic streams on other roads, the starting edges are moved to the upstream intersection and the ending edges are moved to the downstream intersection. The route planning of a vehicle is determined during traffic reconstruction.

After re-assigning the starting edge and ending edge of each route, duplicate traffic counts can occur. For example, a vehicle traveling through intersection IV from northbound can also travel through intersections I, II, and III, contributing to the northbound count for all four intersections. To avoid duplicated counts, we consider the coordination of traffic flows among adjacent intersections and refine the number of routes for ensuring matching turning counts in simulation and actual data.

We use Geoffrey E. Havers Statistic (GEH), a widely used metric to assess the similarity between simulated traffic flow and real-world traffic flow:

$$\text{GEH} = \sqrt{\frac{2(M - C)^2}{M + C}}, \tag{6}$$

where $M$ and $C$ represent the turning counts of simulated traffic flow and observed traffic flow, respectively. In transportation engineering, it is generally accepted that simulated traffic resembles real-world traffic when GEH <5 (Timothy and Marzenna, 2005; El Esawey and Sayed, 2011). We compute the average GEH using all turning counts at all six intersections. The resulting values are 1.97, 1.49, 1.84, 2.34, 1.87, and 2.19, indicating our simulation's high fidelity.

*4.1.2. Mixed traffic generation.* To create mixed traffic, a newly spawned vehicle will be randomly assigned to be either RV or HV according to a pre-specified RV penetration rate. For HV, the longitudinal acceleration is computed using intelligent driver model (IDM) (Treiber et al., 2000). For RV, when it is outside the control zone, IDM is used to determine the longitudinal acceleration; when it is inside the control zone, the high-level decisions Stop/Go are determined by the RL policy, while the low-level longitudinal acceleration is determined by the formulas introduced in Section 3.2.1.

### 4.2. Experiment set-up

Our evaluation metric is the average waiting time (Zhang et al., 2020; Gregurić et al., 2020) of all vehicles. We define the waiting time of each vehicle as the total consecutive time it remains still in the control zone. The average waiting time for a moving direction is the mean of the waiting times of all vehicles in that direction, while the average waiting time for an intersection is the mean of the waiting times of all vehicles at the intersection.

We evaluate our method by comparing it to four baselines: (1) **TL**: the traffic signal program deployed in the city of Colorado Spring, CO; (2) **NoTL**: no traffic lights; (3) **Yan**: the state-of-the-art RL traffic controller with 100% RV penetration rate (Yan and Wu, 2021);[3] and (4) **Yang**: the state-of-the-art CAV control method for unsignalized intersections (Yang and Oguchi, 2020). Our RVs are trained at intersections I, II, III, and IV. We evaluate RVs' performance at all six intersections, including the unseen, three-way intersection VI. All intersections are described in Table 2. Furthermore, we evaluate our method on three manually-generated intersections with different topologies (3-Lane, 4-Lane, and 7-Lane) and traffic demands. The details of these scenarios are introduced in Table 4.

### 4.3. Overall performance

Table 3 shows the main results measured with reduced average waiting time in percentage at intersections I, II, III, and IV. We test RV penetration rates from 20% to 100%, conducting 10 experiments at each rate and reporting the averaged results. Each experiment lasts 1000 steps (1000 s in simulation) and is repeated 100 times. The location and behaviors of HVs are stochastic. The performance of our method varies at different intersections. With only 20% RVs, our method can surpass the traffic light control at intersection II where we perform best. At other intersections, our method with 60% can outperform the traffic signal control baseline. An example comparing our approach with using traffic lights on all moving directions at intersection I is shown in Figure 4. In the absence of traffic lights and with 100% RVs, we can achieve up to 89% reduction in average waiting time. These findings show that our approach can scale to various RV penetration rates while efficiently coordinating mixed traffic.

### 4.4. Individual intersection performance

In Figure 5, we show the detailed performance of our method at intersection I. The results include two parts. The LEFT sub-figure reports the influence of different RV penetration rates on the average waiting time. The RIGHT sub-figure displays the zoomed-in version of the LEFT sub-figure by excluding the NoTL and Yan methods because of their subpar performance. We can see that the average waiting time continuously reduces when the RV penetration rate increases from 20% to 100%. In general, our method starts to outperform TL and Yang when the RV penetration rate is 60% or higher.

We further show traffic congestion levels of intersection I in Figure 4. The congestion level $CL$ is defined as $CL = \min (AWT/\text{Threshold}, 1.0)$, where $AWT$ is the average waiting time of all vehicles in a specific direction, which is normalized by the median value of the average waiting time of the baseline TL at the intersection, that is, Threshold. The Threshold is 46.5 for intersection I. As a result, traffic controlled via our method achieves much lower congestion levels than TL. In addition, our method can flexibly coordinate conflicting moving directions based on varied traffic conditions, which is different from TL that employs fixed-phase coordination.

Figures 6–8 show the detailed performance of our method at intersections II, III, and IV, respectively. Similar results are observed: Our method significantly overtakes Yang and TL with 20%, 70%, and 50% RV penetration rates.

### 4.5. Blackout

To demonstrate our approach's robustness, we simulate blackout events in which all traffic signals are suddenly off. Figure 9 shows the results of no RV and 50% RVs. In case of no RV, a gridlock will form at the intersection within 15 min after the traffic lights are off (starting from the 5th minute). In contrast, with 50% RVs, no congestion is observed. See Figure 9 for demonstrations. More results are shown in Figure 10. The blackouts occur at the 100th step. Without RVs, the absence of traffic lights leads to significant increases in average waiting time due to gradually formed congestion. However, with the presence of 50% RVs, the average waiting time remains stable during the blackouts. Essentially, our method enables RVs to act as "self-organized traffic lights," which effectively coordinate traffic at the intersection.

We also examine the impact of sudden RV rate drops on our approach. The sudden drops could occur due to unstable V2V communication, software failures, humans taking over the control, and more. The "offline" RVs are simulated using the IDM model (Treiber et al., 2000). The results are shown in Figure 11. All drops occur at the 100th step. Our method has significantly reduced the waiting time compared to NoTL. The trials using NoTL display exponential increases in average waiting time, indicating the form of traffic deadlocks. In contrast, at intersection II, our approach effectively stabilizes the system and avoids congestion by maintaining a low average waiting time. At the other three intersections, our method stabilizes the system when the RV penetration rate reaches 90%, with no significant increase in waiting time. Considering other RV penetration rates, we observe an overall increase in average waiting time as incoming traffic flows continue. We analyze the slope of the average waiting time curves to identify the rising trend of congestion levels. As shown in Figure 12, if the slope approaches 1, there is a higher likelihood of deadlock. However, in scenarios where our method is deployed, the rising trend plateaus below 0.4 after 2000 simulation steps show effective congestion management. In comparison, NoTL's AWT trend approaches nearly 0.8, hinting severe congestion.

**Table 2.** Real-world intersections used in training and testing (see I, II, III, and IV in Figure 1 and V and VI in Figure 14). Actual traffic demand is provided. We also list the number of non-empty lanes since some lanes do not show traveling vehicles.

| Intersection | Num. incoming lanes | Num. non-empty lanes | Traffic demand (v/h * lane) |
|---|---|---|---|
| I | 21 | 19 | 1157 |
| II | 19 | 18 | 1089 |
| III | 18 | 17 | 928 |
| IV | 16 | 14 | 789 |
| V | 24 | 24 | 987 |
| VI | 10 | 10 | 1115 |

**Table 3.** Reduced average waiting time in percentage at each intersection under different RV penetration rates. Our method outperforms traffic signals in all cases when the RV penetration rate ≥ 70%. More time is saved with higher RV penetration rates. Compared to scenarios without traffic lights, our method can achieve up to 89% reduction in average waiting time with 100% RVs.

| | Reduced average waiting time (%) | | | | | | |
|---|---|---|---|---|---|---|---|
| | Compared to TL (%) | | | | | | Compared to NoTL (%) |
| Intersection I | −15.04 | 0.34 | 25.98 | 32.60 | 40.75 | 42.01 | 89.91 |
| Intersection II | 41.78 | 48.77 | 48.90 | 51.84 | 53.46 | 52.80 | 61.46 |
| Intersection III | −40.53 | −2.83 | 2.75 | 13.39 | 10.60 | 22.33 | 66.72 |
| Intersection IV | −15.91 | 8.20 | 9.70 | 44.47 | 52.19 | 66.63 | 81.73 |
| RV rate | 50 | 60 | 70 | 80 | 90 | 100 | 100 |

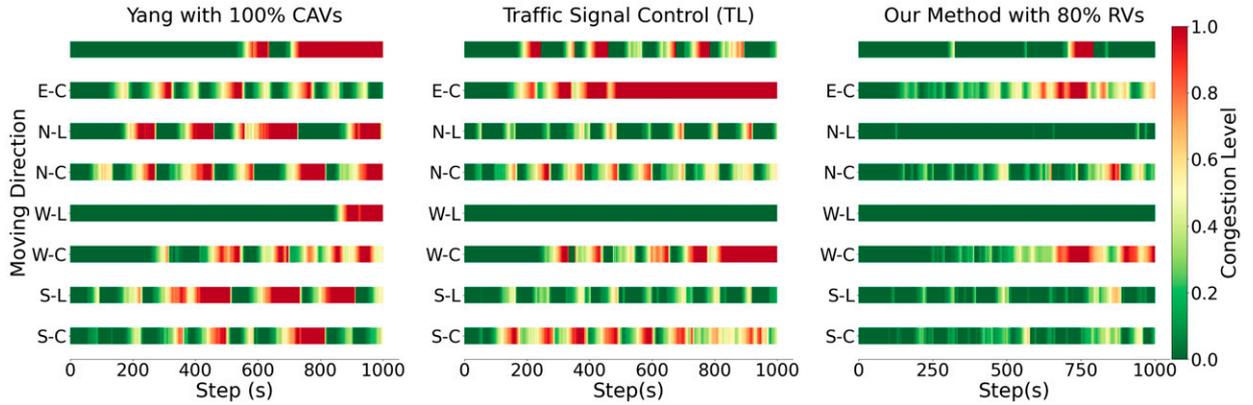

**Figure 4.** Traffic congestion levels at intersection I under different control mechanisms. Our approach with 80% RVs consistently achieves lower congestion levels than Yang and TL. Unlike Yang and TL, which control intersection traffic using fixed phases, our method learns to use adaptive phases in control.

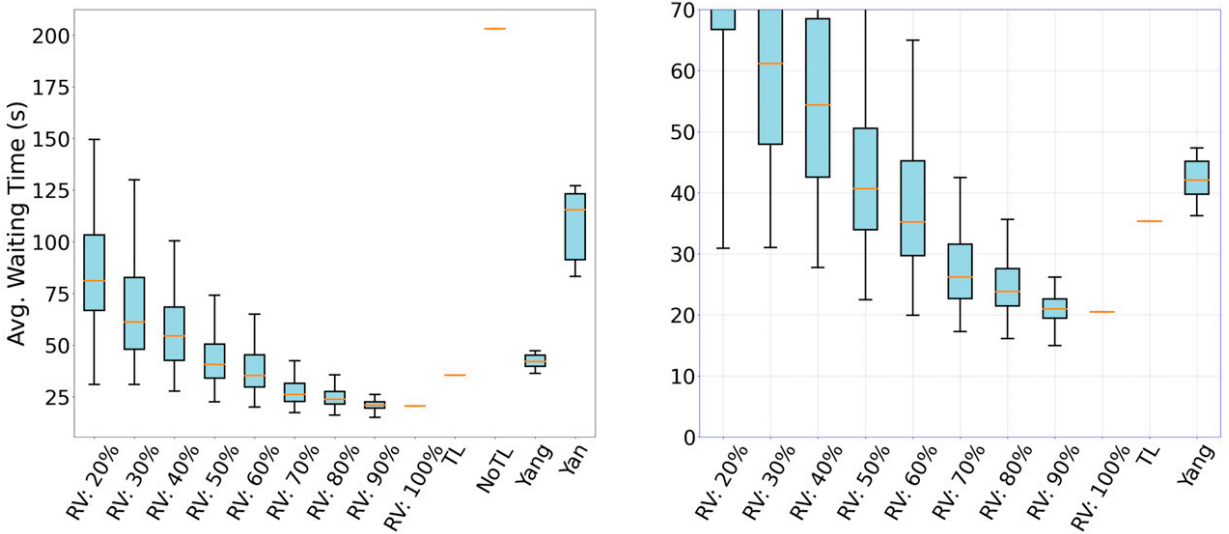

**Figure 5.** The overall results in average waiting time at intersection I. The RIGHT sub-figure displays zoomed-in version of the LEFT sub-figure, excluding the NoTL and Yan methods. When the RV penetration rate reaches or exceeds 60%, our method consistently outperforms the other four baselines.

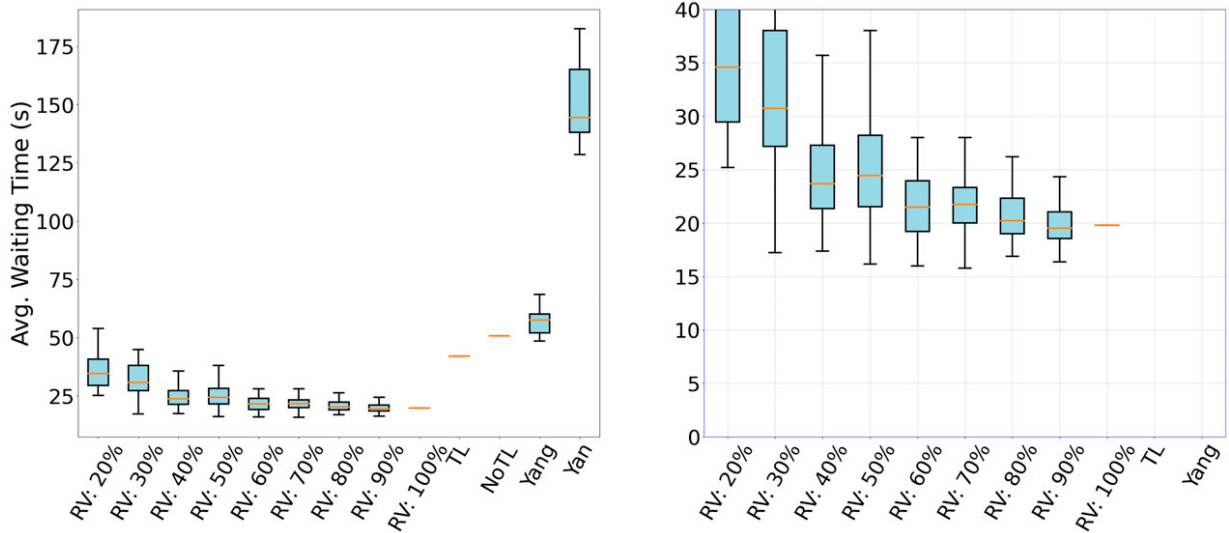

**Figure 6.** The overall results in average waiting time at intersection II. The RIGHT sub-figure displays zoomed-in version of the LEFT sub-figure, excluding NoTL and Yan. With 20% or more RVs, our method consistently outperforms all other baselines.

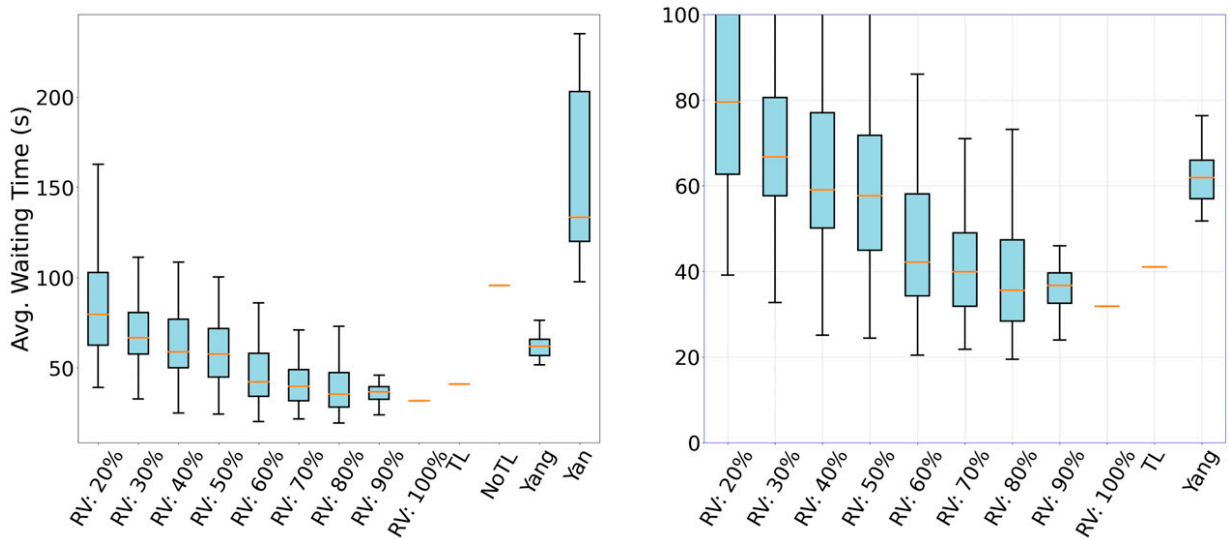

**Figure 7.** The overall results in average waiting time at intersection III. The RIGHT sub-figure displays zoomed-in version of the LEFT sub-figure, excluding NoTL and Yan, which perform significantly worse than our method. Our method begins to outperform TL and Yang when the RV penetration rate reaches 60–70% or higher.

### 4.6. Traffic demands and congestion

We further analyze the relationship of traffic demands and congestion. The results using intersection I as the testbed are shown in Figure 13. By increasing the traffic demand from 150 v/h to 300 v/h with no traffic lights and no RVs, we observe congestion starting to form at 200 v/h, indicated by a low average speed of all vehicles at the intersection (we define congestion when average vehicle speed < 1 m/s.

In contrast, with the actual traffic demand 700 v/h, congestion does not form with just 5% RVs in mixed traffic controlled by our algorithm. Figure 13 also demonstrates that the minimum RV penetration rate required to avoid congestion under the real-world traffic demand is 5%.

### 4.7. Generalization

To evaluate the generalizability of our approach, we first test it on two previously unseen real-world intersections shown in

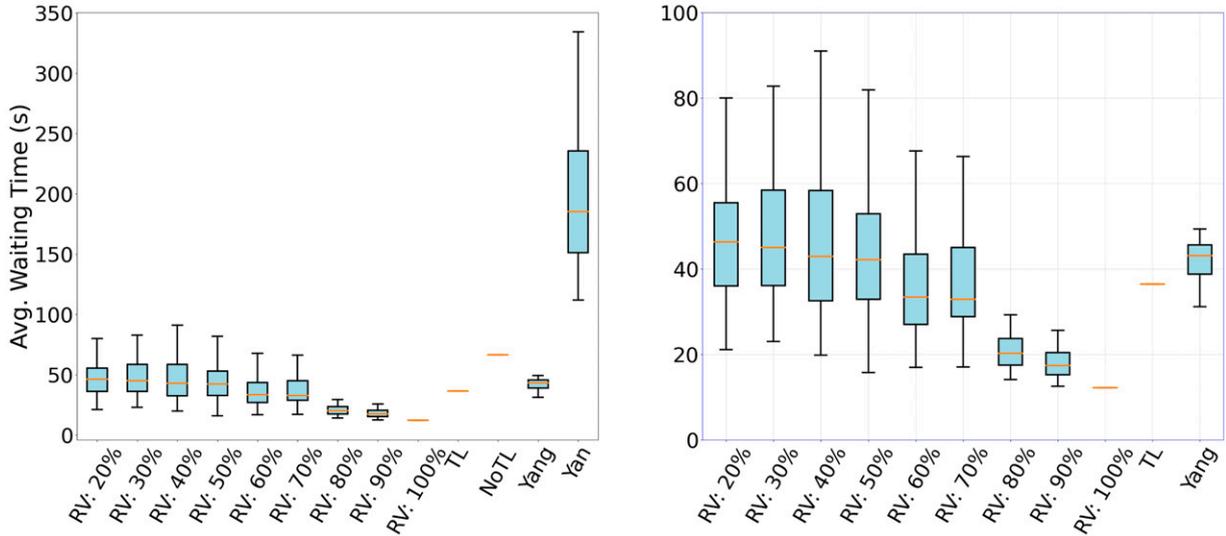

**Figure 8.** The overall results in average waiting time at intersection IV. The RIGHT sub-figure displays zoomed-in version of the LEFT sub-figure, excluding NoTL and Yan. Our method with 60% RVs or more outperforms all four baselines.

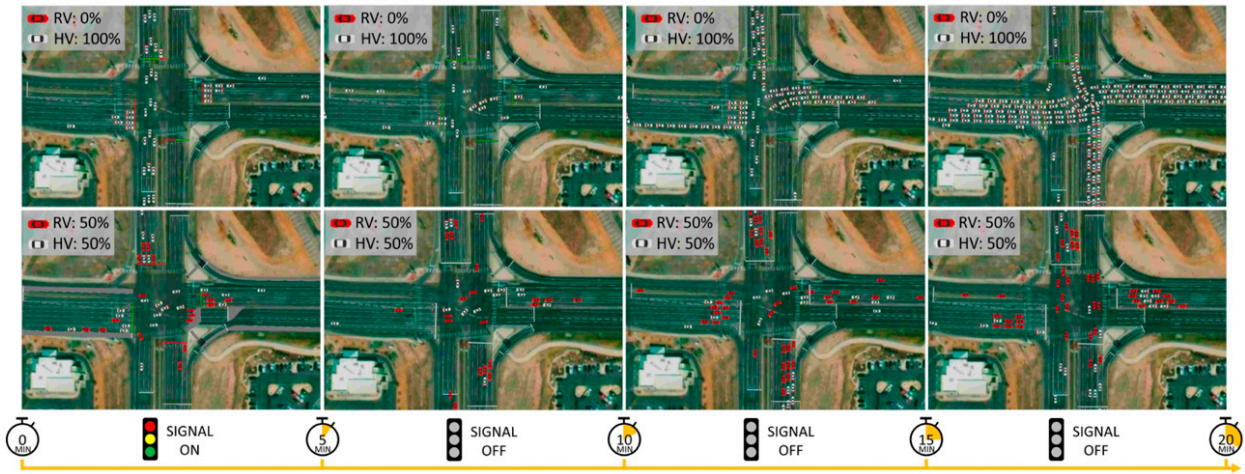

**Figure 9.** Comparison between traffic conditions with and without RVs during a blackout event at intersection I. The blackout event occurs at the 5-min mark. Congestion forms rapidly within 15 min in traffic without RVs. Conversely, traffic regulated with 50% RVs does not result in congestion.

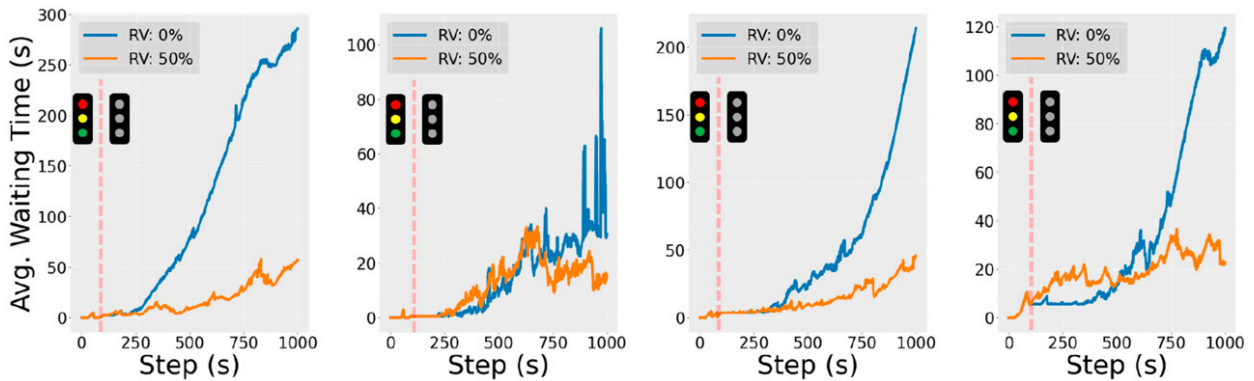

**Figure 10.** Blackout experiments. We simulate blackout events (traffic signals off) at intersections I, II, III, and IV (from left to right) since the 100th step. Without any RV, a gridlock will form at the intersection causing the average waiting time of all vehicles to increase rapidly. In contrast, with 50% RVs, no gridlock appears and the waiting time of all vehicles remains low and stable.

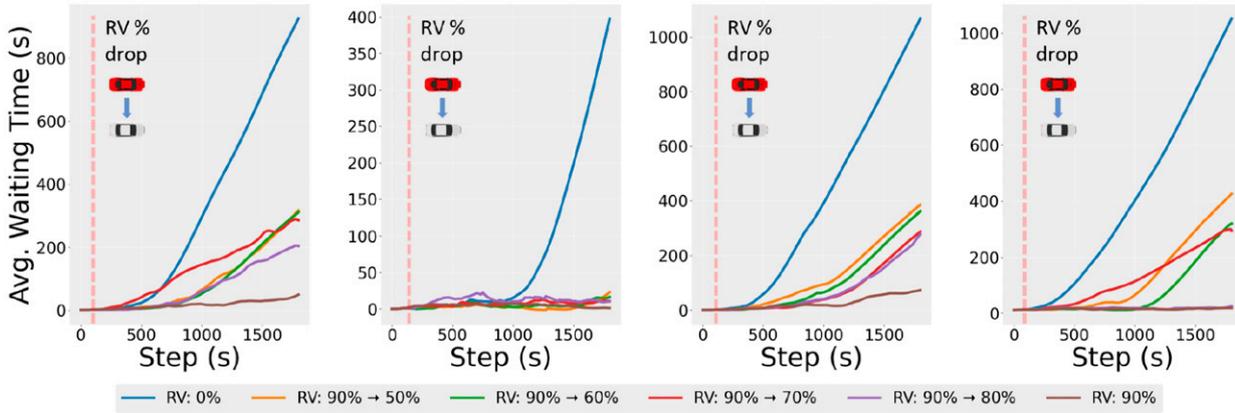

**Figure 11.** Our method ensures stable and uncongested traffic, even when the RV penetration rate abruptly drops. The sub-figures from left to right correspond to intersections I, II, III, and IV. NoTL displays exponential increases in average waiting time, indicative of traffic deadlocks.

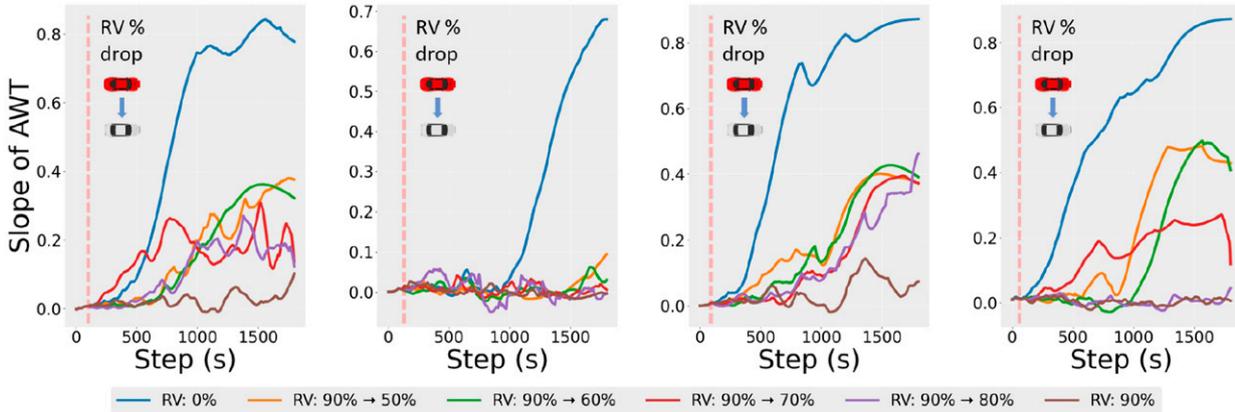

**Figure 12.** The slope of average waiting time (AWT) when the RV penetration rate suddenly drops. The sub-figures from left to right correspond to intersections I, II, III, and IV. The slope of AWT successfully plateaus when our method is activated, indicating effective congestion management.

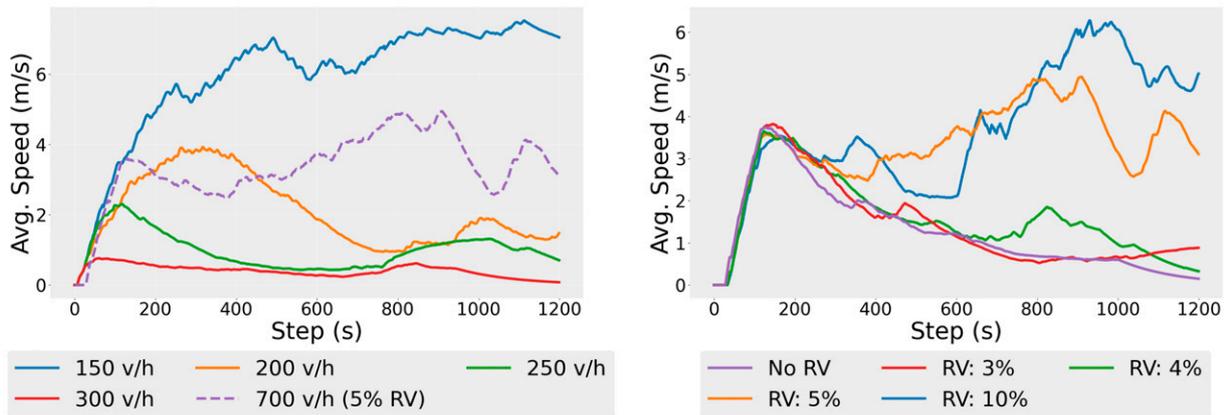

**Figure 13.** LEFT: The solid lines represent no traffic lights and no RVs. The congestion starts to form when the demand is over 200 v/h. The real-world demand denoted using the dashed line, which is about 700 v/h, does not build congestion because 5% RVs are deployed in traffic. RIGHT: Analyzing the influence of low RV penetration rates on traffic. As a result, 5% is the minimum to prevent congestion. For both figures, the study subject is intersection I.

Figure 14, one of which is a three-way intersection. We apply our policy directly to the unseen four-way intersection without refinement, and it functions well. As shown in Figure 15(a), our policy can achieve comparable performance to conventional traffic signal baseline when the RV penetration rate is 60% or higher. Our policy is also deployed without refinement at the unseen three-way intersection, which requires coordination among only four directions: S-C, S-L, W-L, and N-C. Our policy adapts to this case by setting the input values of other directions to zero. Despite never encountering this topology and the corresponding traffic demand, our policy coordinates traffic well. As shown in Figure 15(b), our approach outperforms the traffic light baseline when RV penetration is at 40% or higher. At 90% RV penetration, our method reduces the average waiting time by approximately 62.5% compared to using traffic lights. These results demonstrate the excellent generalizablility of our approach.

We further investigate the generalizability by testing our approach at three manually-created intersections, each with a distinct topology and traffic demand. The intersections have different numbers of incoming lanes per direction, allowing us to simulate a wide range of traffic scenarios, as shown in Figure 16. Traffic flows are generated using the profiles detailed in Table 4. The results, presented in Figure 17, demonstrate that our method can effectively coordinate traffic at these unseen environments as well. With approximately 70% of RVs, our approach outperforms traditional traffic light control.

### 4.8. Evaluation of reward function

Designing effective rewards for controlling mixed traffic at complex intersections is an open problem to date. Varied intersection topology, conflicting traffic streams, and the use of real-world traffic data that can lead to unpredictable and unstable inflow/outflow all complicate the task. To address conflicts within intersections and prevent traffic congestions, our insight is to consider the impact of each RV's actions on traffic flow in its own direction, while penalizing conflicting decisions among RVs.

The designed reward should promptly reflect the severity of congestion and traffic conditions at an intersection. Figure 18 illustrates the traffic conditions during NoTL (no traffic lights and 100% HVs) and average vehicle speed of using our reward and Yan and Wu (2021)'s reward. The average vehicle speed is used to indicate congestion severity. We consider an intersection congested if the average speed is < 1 m/s. In Figure 18, congestion arises when there are no traffic lights and 100% HVs, resulting in a rapid decrease in the average speed of all vehicles (orange curve). Our reward (blue curve) promptly captures this trend, making it an effective indicator for the learning process. In contrast, the reward of the state-of-the-art method by Yan and Wu (2021) fails to do so.

The reward function by Yan and Wu (2021) takes the format $R_{Y\,an} = \texttt{outflow}(s_t, s_{t+1}) - \texttt{collision}(s_t, s_{t+1})$, where $\texttt{outflow}(s_t, s_{t+1})$ denotes the number of vehicles exiting the network from $t$ to $t+1$, and $\texttt{collision}(s_t, s_{t+1})$ is the number of collisions in the network from $t$ to $t+1$. We record this reward during the evaluation of NoTL to analyze its characteristics. As anticipated, the absence of traffic lights leads to intersection congestion, evidenced by the average speed of all vehicles rapidly approaching zero. However, Yan's reward fails to capture changes in traffic conditions promptly. This is due to the outflow of a network being a delayed indicator; congestion within the intersection does not prevent previously cleared vehicles from contributing to the outflow. Consequently, the delayed reward hinders the learning process, with episodes often terminating due to congestion before the reward can manifest it. In contrast, our reward function captures the immediate impact of traffic conditions by incorporating the waiting time of traffic streams. This provides the RL agent with an explicit optimization objective.

### 4.9. Evaluation of conflict resolution

Our algorithm and reward design aim to minimize potential conflicts within the intersection, which not only improves the effectiveness of RL training, but also implicitly improves the road safety inside the intersection. To evaluate the effectiveness of our reward function in avoiding conflict, we calculate the conflict rate as the ratio of conflicts to the number of RVs within the control zone. As shown in Figure 3 MIDDLE, the number of conflicts decreases as training progresses and stabilizes at a low level. This indicates that the policy successfully learns to prevent conflicting movements. In Figure 3 RIGHT, the conflict rate for 80% RVs converges around 4%, while the conflict rate for 100% RVs is less than 3% after 500 steps during evaluation. These results demonstrate that our trained RL policy successfully learns to avoid conflicts, thereby enhancing traffic safety.

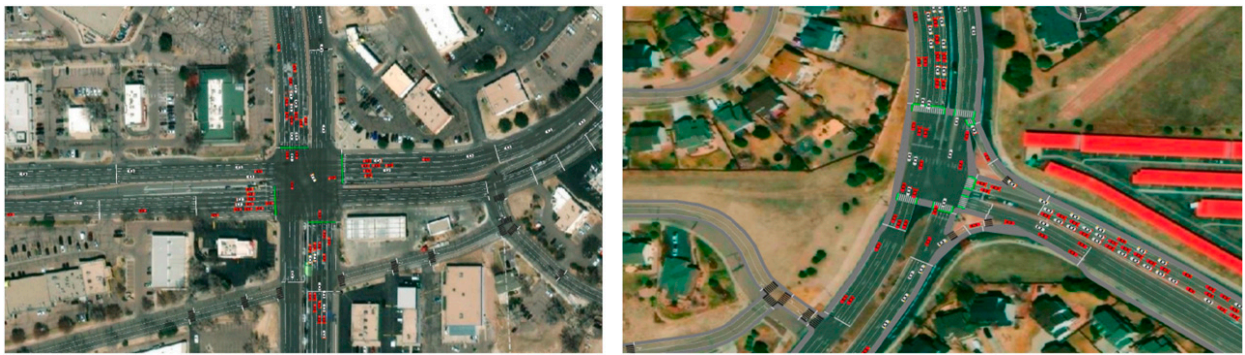

**Figure 14.** Two unseen real-world intersections used in our test. LEFT: Intersection V, four-leg. RIGHT: Intersection VI, three-leg.

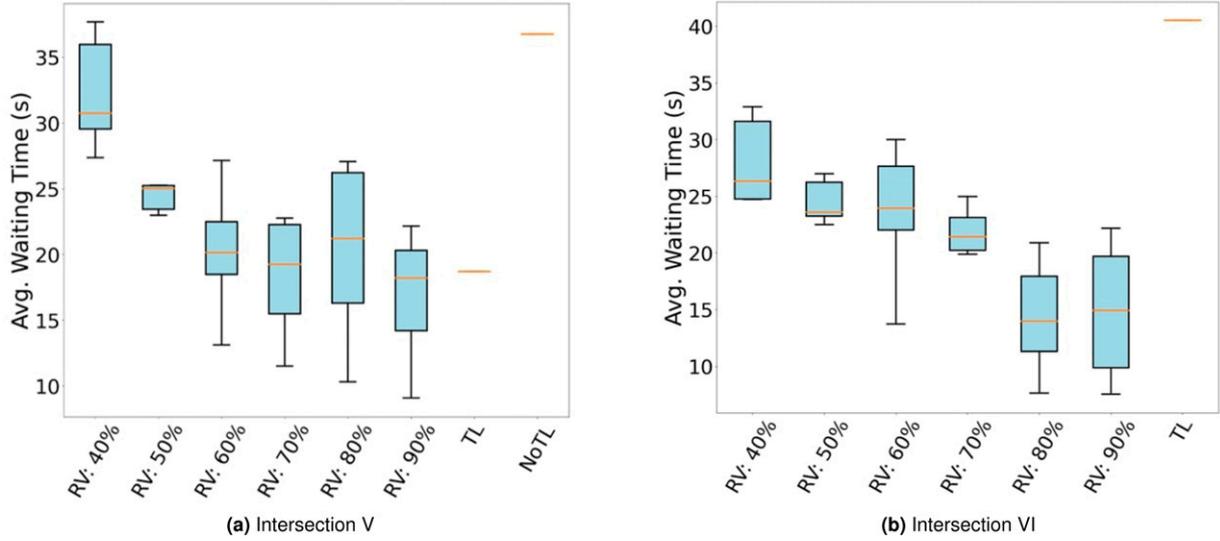

**Figure 15.** Overall results in the average waiting time at the intersection V and VI. Starting from 60–70% RVs, our method outperforms the traffic lights (TL) baseline at the intersection V. Additionally, our approach starts to outperform the TL baseline when RVs are 40% or more at the intersection VI. (a) Intersection V. (b) Intersection VI.

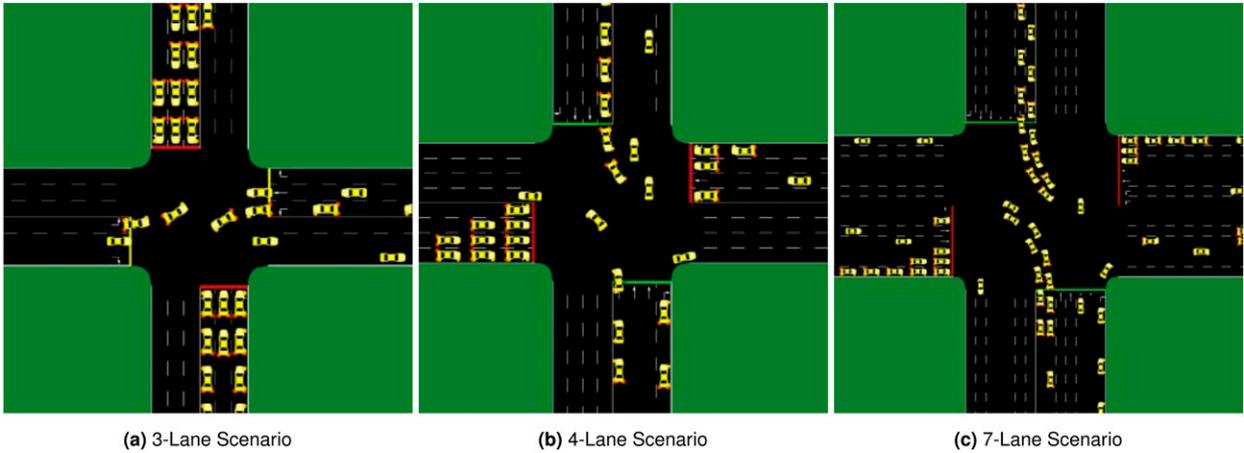

**Figure 16.** Three manually-created intersections in SUMO to further investigate our method's generalization capability. Each intersection has different topologies and traffic demands. (a) 3-lane scenario. (b) 4-lane scenario. (c) 7-lane scenario.

**Table 4.** The traffic demand profiles of the manually-generated intersections.

| Scenario | Num. incoming lanes | Traffic demand (v/h * lane) |
|---|---|---|
| 3-Lane | 12 | 550 |
| 4-Lane | 16 | 412 |
| 7-Lane | 28 | 235 |

To justify the effectiveness and necessity of conflict resolution mechanism in training, we conduct ablation studies. As depicted in Figure 19, the absence of the conflict resolution mechanism (PPO w/o Mech, SAC w/o Mech and R. DQN w/o Mech) leads to no reduction in waiting time, indicating ineffective learning. In contrast, with the conflict resolution mechanism, all RL algorithms demonstrate rapid learning and convergence.

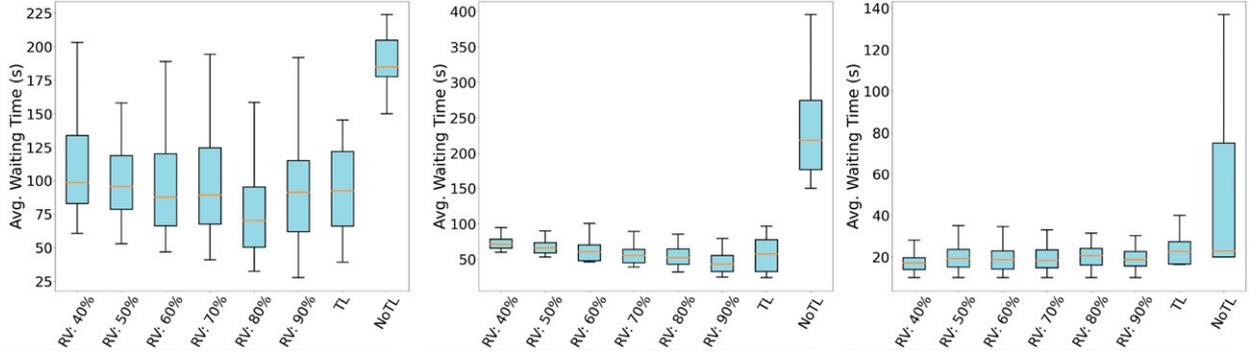

**Figure 17.** From left to right, our results at manually-generated 3-lane, 4-lane, and 7-lane intersections shown in Figure 16. Our method successfully regulates traffic in these unseen intersections without fine-tuning.

## 4.10. Robustness in potential real-world deployment

Our method is designed for decentralized execution. As we discussed in Section 3.2.2, the input to our policy requires both V2V information and local perception. The inference process is described in Alg. 1.

---

**Algorithm 1** Real-world Inference for RV $i$

---

1: System starts and loads policy model
2: **for** $t = 1, 2, \ldots,$ **do**
3:   Get state information: localization result and perception result.
4:   Compute distance $d^t$ between ego vehicle and the intersection.
5:   **if** $d^t < 30\,\mathrm{m}$ **then**
6:     Receive shared information from other RVs: $\{(\text{vehicle id } k, \text{position } \mathrm{pos}_k^{t-1}, \text{ego waiting time } {}^{\mathrm{ego}}w_k^{t-1}, \text{queue length } {}^{\mathrm{ego}}l_k^{t-1})\}$
7:     Aggregate the information according to intersection traffic streams, as described in Sec. 3.2.3.
8:     Reconstruct the observation, as defined in Sec. 3.2.2 and feed it into the policy.
9:     Get policy output and execute the decision.
10:     Send its own ego state $(i, \mathrm{pos}_i^t, {}^{\mathrm{ego}}w_i^t, {}^{\mathrm{ego}}l_i^t)$ to other RVs by V2V communication if it stops.
11:     Package delivery from source RV to all target RVs utilizes existing communication protocols.
12:   **else**
13:     Skip, because the ego RV $i$ is outside the control zone.
14:   **end if**
15: **end for**
16: System shutdown

---

Our method can be potentially applied to real-world RVs which are equipped with V2V capabilities and self-driving software. However, conducting real-world experiments for mixed traffic control can be expensive. Therefore, we validate our approach through simulation. In particular, whether the decision-making of the RV described in Alg. 1 is robust to traffic conditions estimated from real-world observations. As discussed in Section 3.2.2, RL observations are estimated from local perception and V2V communication using technologies such as LTE, WiMax, or Bluetooth. Due to the variability in connectivity quality and range, multi-hop communication and potential packet errors or losses are inevitable (Vegni and Little, 2010). Hence, it is crucial to assess our method under different communication conditions. To achieve this, we simulate two types of connectivity protocols:

- **Long-range connectivity** (e.g., **LTE and WiMax**): In this protocol, vehicles within 150 m can directly communicate and exchange information with each other using a single hop, without the need for intermediary nodes. See Figure 20 LEFT.
- **Short-range connectivity** (e.g., **Bluetooth**): This protocol uses short-range communication techniques, allowing vehicles to communicate around an intersection with up to three hops, each spanning up to 50 m. The communication process involves four steps: (1) Vehicles in the same direction form clusters, with the leading vehicle designated as the master node; (2) each vehicle transmits its local information to the master node of its cluster; (3) master nodes exchange gathered information with master nodes of other clusters; and (4) the master node distributes the information from other directions back to the slave nodes within its cluster. See Figure 20 RIGHT.

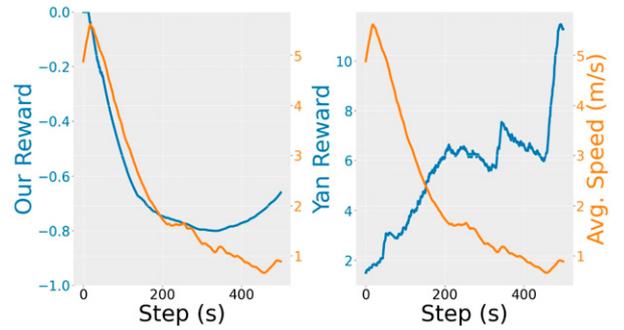

**Figure 18.** When congestion occurs during the NoTL setup, the average speed within the intersection decreases to nearly zero (orange curve). Our reward can promptly respond to congestion, indicated by the decreasing negative reward curve in the LEFT sub-figure. In contrast, the reward from Yan and Wu (2021) incorrectly increases as congestion worsens, as shown in the RIGHT sub-figure.

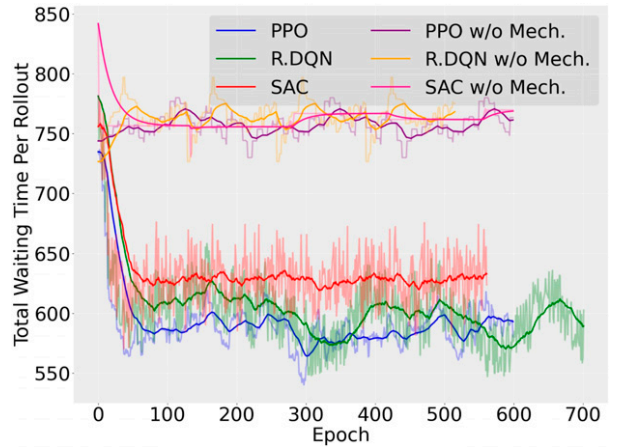

**Figure 19.** Visualization of the cumulative waiting time per rollout when training various RL algorithms, such as SAC, PPO, R. DQN (Rainbow DQN), R. DQN w/o Mech (Rainbow DQN without conflict resolution mechanism), SAC w/o Mech. and PPO w/o Mech. PPO and Rainbow DQN demonstrate similar convergence. However, without our conflict resolution mechanism (Mech), all algorithms fail to learn, indicated by a high waiting time after 500 epochs of training.

Simulating these two distinct connectivity protocols allows us to evaluate the robustness and efficiency of our method. The long-range connectivity ensures minimal latency through direct communication, while the short-range connectivity assesses performance in a more complex, multi-hop, high-latency communication environment. As the message to be shared among vehicles is brief and can be swiftly exchanged, we assume that the communication process between vehicles is rapid, with all communication completed within 1 s, including both one-hop and multi-hop connectivity. This aligns with the fact that our proposed policy operates at 1 Hz.

In addition to connectivity range, the reliability of connectivity is another critical factor in real-world wireless communication. For example, the IEEE 802.11a standard specifies a maximum tolerable package error rate (PER) of 10% (IEEE Standards Association, 1999). Despite the integration of sophisticated network protocols equipped with various error mitigation mechanisms, it is essential to evaluate the impact of connectivity errors on our system. Therefore, we conduct multiple experiments to simulate various package error rates, ranging from 1% to 20%. In these simulations, we assume that each V2V connection operates independently, with an equal chance of encountering connectivity errors. If a connectivity error occurs, the shared information becomes inaccessible to the affected receivers. As a result, the probability of successfully receiving shared information can be calculated as $P_{\text{info}} = (1 - \text{PER})^{\#\text{hops}}$, where PER denotes the package error rate, and #hops denotes the number of hops on the connectivity link. To examine the impact of different connectivity conditions on our estimated traffic, we calculate the relative estimation error (%) as follows:

$$\text{estimation error} = \left| \frac{\text{actual value} - \text{estimated value}}{\text{actual value}} \right| \times 100\%.$$

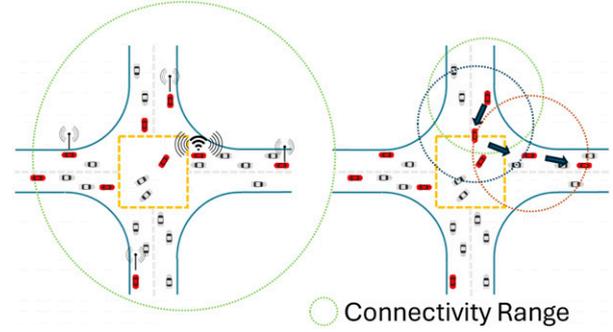

**Figure 20.** Two distinct real-world deployment setups, long-range connectivity and short-range connectivity. LEFT: Long-range connectivity enables seamless broadcasting of vehicle information, accessible to all nearby vehicles around the intersection. RIGHT: With short-range connectivity, messages necessitate up to three hops before reaching their destinations.

This calculation is performed for both estimated queue length and waiting time. The estimation is obtained from the information received from the V2V communication, while the actual value is determined using the ground-truth traffic condition. Based on the relative estimation error visualized in Figures 21(a) and (b), we observe that the error increases with #hops and packet error rate (PER). This leads to increased observation uncertainty for the policy during experiments. For instance, when PER reaches 20% in short-range connectivity, the estimation error of queue length exceeds 20%. This significant error makes decentralized traffic coordination more challenging. However, as shown in Figure 22, our method demonstrates robustness to varying V2V conditions, including multi-hop communication (one-hop and three-hop) and different PER assumptions (1%, 5%, 10%, 15%, and 20%). Remarkably, our method maintains robust performance even under extreme conditions, such as three-hops and 20% PER. When the probability of connectivity is low, such as one-hop communication with 1% PER, our method achieves performance similar to the RL policy with precise observation. However, as observation uncertainty increases, the coordination performance diminishes but is still acceptable. These experiments highlight the robustness of our method and showcase its potential for real-world deployment.

### 4.11. Extension to right-turn traffic

Our simulation is constructed using Geographic Information System (GIS) data sourced from the U.S., a country with right-hand traffic norms. However, traffic regulations vary across nations; for instance, in left-hand traffic countries, right-turn traffic requires control. Our method maintains adaptability to accommodate diverse scenarios with varying traffic rules. By adjusting traffic streams and conflict-free movements, our methods can seamlessly adhere to specific requirements, for example, incorporating and regulating the right-turn traffic.

To showcase this adaptability, we have retrained the policy to include right turns, with 70% RVs and tested it at intersection V. This particular intersection does not have a right-turn ramp, meaning that vehicles from all four incoming directions can enter the intersection. As indicated in Table 5, we report the average waiting time of all vehicles across all 12 moving directions. The new policy, now accommodating right turns, achieves performance comparable to our original policy. Notably, the average waiting time of the right-turn traffic is significantly shorter compared to other directions, which also demonstrates that the right-turn traffic has minimal impact within the traffic scenarios analyzed in this study. This experiment illustrates the flexibility of our method and its potential applicability across various traffic rules.

## 5. Discussion

While our approach demonstrates promising results in large-scale mixed traffic control at complex intersections, we acknowledge several inherent limitations that are worth further exploration.

First, our current learning framework lacks a hierarchical structure, which could be supplemented by low-level sub-policies such as longitudinal and lateral de/accelerations, as well as lane changes. This deficiency can hinder the granularity and responsiveness of traffic regulation, potentially compromising the efficiency and safety of both RVs and HVs. For example, when the penetration rate of RVs is low, the absence of lane change functionality in RVs can result in congestion and deadlock, as HVs can simply bypass RV regulations by changing lanes frequently. To address this issue, RVs should be able to change lanes or decelerate to influence nearby HVs, encouraging them to comply with RV's regulations for more efficient traffic. Additionally, RVs should actively regulate traffic flow within intersections, rather than simply acting as mobile traffic lights. While this concept offers greater regulation potential, it also presents significant challenges in complex intersectional traffic scenarios.

Second, our method employs a relatively rigid conflict resolution mechanism, in terms of a pre-defined if-else rule that restricts vehicle movement at intersections. Though the constrained exploration space is beneficial for training efficiency and safety (Alshiekh et al., 2018), it is still possible that the optimal solution lies beyond the exploration boundaries set by this mechanism. Therefore, future exploration could consider relaxing these constraints, for example using real-world demonstration data (Leung et al., 2023), which could improve overall traffic efficiency without compromising safety.

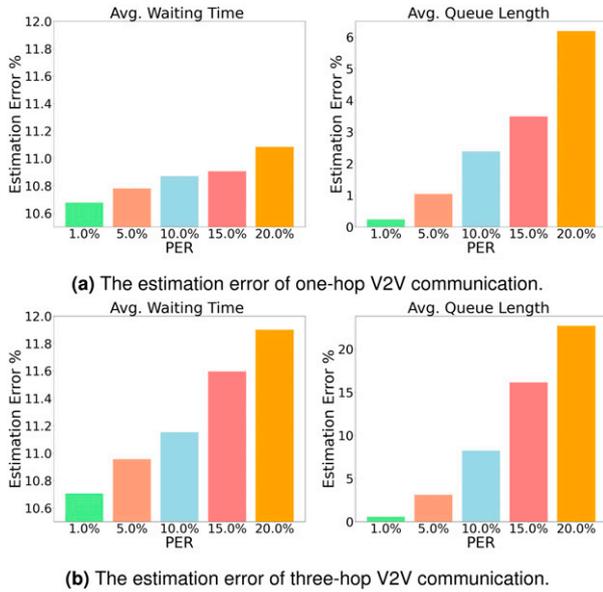

**Figure 21.** The relative error of traffic condition estimation under different V2V communication protocols and PER (package error rate) levels. (a) The estimation error of one-hop V2V communication. (b) The estimation error of three-hop V2V communication.

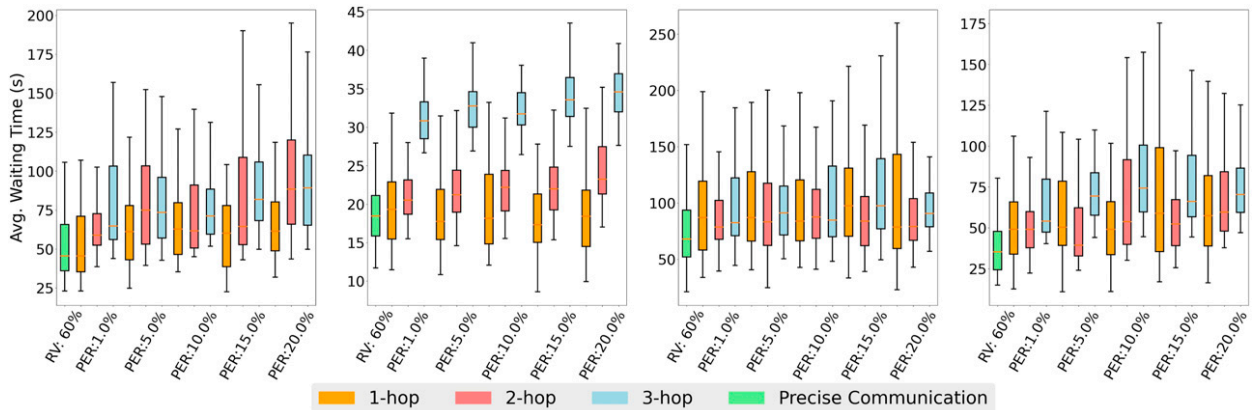

**Figure 22.** From left to right, evaluation of our method in simulated V2V communication experiments with 60% RVs at intersections I, II, III, and IV. The results demonstrate that our method can effectively address the communication uncertainty arising from multi-hop communication and low-quality connection.

**Table 5.** Result of right-turning experiments. We report the average waiting time of all vehicles along each moving direction.

| Directions | S-C | S-L | S-R | W-C | W-L | W-R | N-C | N-L | N-R | E-C | E-L | E-R | All |
|---|---|---|---|---|---|---|---|---|---|---|---|---|---|
| Right-turn-enabled | 27.4 | 12.0 | 1.79 | 20.9 | 11.0 | 0.23 | 31.7 | 14.0 | 0.77 | 19.8 | 9.67 | 0.67 | 18.3 |
| Baseline | 15.4 | 18.7 | N/A | 16.6 | 18.8 | N/A | 21.1 | 16.8 | N/A | 26.8 | 15.8 | N/A | 19.9 |

Third, our traffic status encoder is currently designed for intersections with a maximum of four ways, and it must be expanded to accommodate other topologies, such as multiway intersections and roundabouts. Addressing this limitation is crucial to broaden the usability and adaptability of the proposed method. One possible solution is to use a graph convolutional network (GCN) to model the intersection. However, existing GCN methods like Bai et al. (2020) only model the topology of the intersection and cannot capture the complex geometric details inside the intersection that is necessary for real-world intersection traffic regulation.

Fourth, we use IDM (Treiber et al., 2000) to simulate interactions between RVs and HVs. However, IDM can be replaced by other advanced methods, such as learning-based simulations for autonomous driving and traffic (Guo et al., 2024), to capture the accurate interaction between HVs and RVs. Nevertheless, these learning-based models may sometimes be unstable, leading to erroneous outputs that could negatively impact the RL agent. Additionally, most of the traffic data used in these methods are based on HVs' regular trajectories in stable traffic, lacking the active responses of HVs to mixed traffic conditions. Therefore, an important next step would be to collect real-world traffic data in complex intersections with dynamic and rich HV–RV interaction.

Finally, we currently rely solely on RV information within the control zone to estimate traffic conditions, which may lead to significant errors. Although HVs in mixed traffic cannot be directly controlled like RVs, they can still provide valuable traffic information. For example, by utilizing onboard GPS or Google Map APIs, HVs can measure their ego queue lengths and waiting times in the direction they are moving. This data can enhance the accuracy of traffic condition estimation, resulting in improved traffic regulation performance within our RL framework. However, it also brings additional challenges, including privacy concerns and the need for standardized communication protocols. These challenges can be addressed by implementing privacy-preserving crowdsourcing techniques (Wang et al., 2023a).

## 6. Conclusion

We propose a decentralized RL approach for the control and coordination of mixed traffic at real-world and unsignalized intersections. Our method tackles the most intricate mixed traffic intersection scenarios when compared to existing literature, encompassing diverse intersection capacities, topologies, and fluctuating traffic demands. Our approach encompasses numerous innovative designs tailored for mixed traffic control. To address the diverse range of road topologies and traffic demands, we introduce a generic representation for encoding intersection traffic conditions, ensuring enhanced generalizability. To optimize traffic efficiency and minimize conflicting movements, we devise a conflict-aware reward function specifically for coordinating large-scale mixed traffic at intersections. This function not only serves as a timely incentive for the RL agent but also boosts the training performance. We also introduce a conflict resolution mechanism aimed at preventing conflicts as well as improving training efficiency. Lastly, we provide a high-fidelity traffic simulation reconstructed using real traffic data for robust training and testing.

Our method is the first to control mixed traffic under real-world traffic conditions at complex intersections. Various experiments are conducted to show the effectiveness, robustness, generalizablility, and adaptability of our approach. Detailed analyses are also pursued to justify the design of the components of our method. Our method can serve as an inspiration for mixed traffic control via model-free RL at large-scale and complex scenarios, paving the way for next-generation traffic control strategies.

## Declaration of conflicting interests

The author(s) declared no potential conflicts of interest with respect to the research, authorship, and/or publication of this article.

## Funding

The author(s) disclosed receipt of the following financial support for the research, authorship, and/or publication of this article: Dawei Wang and Jia Pan are partially supported by the Innovation and Technology Commission of the HKSAR Government under the InnoHK initiative, and Research Grant Council (Ref: 17200924), Hong Kong. National Science Foundation; IIS-2153426 and Innovation and Technology Fund; GHP/126/21GD.

## ORCID iD

Jia Pan  https://orcid.org/0000-0001-9003-2054

## Notes

1. https://coloradosprings.gov/.
2. https://sumo.dlr.de/docs/jtrrouter.html.

3. To apply this approach, we extend the network input to the maximum number of incoming lanes to accommodate a real-world intersection.